\documentclass{ifacconf}

\usepackage{amssymb}
\usepackage{amsmath}
\usepackage{graphicx} 
\graphicspath{ {images/} }
\usepackage{listings}
\usepackage{algorithm,algcompatible}
\makeatletter
\def\munderbar#1{\underline{\sbox\tw@{$#1$}\dp\tw@\z@\box\tw@}}

\usepackage{algorithm}
\usepackage{algorithmicx}
\usepackage{algpseudocode}
\usepackage{booktabs}       
\usepackage{url}
\usepackage{lipsum}
\usepackage{bbm}
\usepackage{soul}

\usepackage{natbib}        

\makeatletter
\def\endfigure{\end@float}
\def\endtable{\end@float}
\makeatother

\let\ifacconfcaptionwidth\captionwidth
\usepackage[caption=false]{subfig}
\let\captionwidth\ifacconfcaptionwidth

\newcommand{\R}{\mathbb{R}}
\newcommand{\Z}{\mathbb{Z}}
\newcommand{\x}{\mathbf{x}}

\newcommand{\uu}{\mathbf{u}}
\newcommand{\s}{\mathbf{s}}

\newcommand{\Ss}{\mathcal{S}}
\newcommand{\W}{\mathcal{W}}

\newcommand{\U}{\mathcal{U}}

\newtheorem{theorem}{Theorem}
\newtheorem{problem}{Problem}
\newtheorem{definition}{Definition}
\newtheorem{assumption}{Assumption}

\begin{document}
\begin{frontmatter}

\title{A Formal Safety Characterization of Advanced Driver Assist Systems in the Car-Following Regime with Scenario-Sampling\thanksref{footnoteinfo}} 


\author[First]{Bowen Weng$^{+}$} 
\author[First]{Minghao Zhu$^{+}$} 
\author[First]{Keith Redmill}
\thanks{$^+$These authors contributed equally.}
\address[First]{Department of Electrical and Computer Engineering at Ohio State University, OH, 43210 USA (e-mail: weng.172@osu.edu, zhu.1385@osu.edu, redmill.1@osu.edu).}

\begin{abstract}    
The capability to follow a lead-vehicle and avoid rear-end collisions is one of the most important functionalities for human drivers and various Advanced Driver Assist Systems (ADAS). Existing safety performance justifications of car-following systems either rely on simple concrete scenarios with biased surrogate metrics or require a significantly long driving distance for risk observation and inference. In this paper, we propose a guaranteed unbiased and sampling efficient scenario-based safety evaluation framework inspired by previous work on $\epsilon\delta$-almost safe set quantification. The proposal characterizes the complete safety performance of the test subject in the car-following regime. The performance of the proposed method is also demonstrated in challenging cases including some widely adopted car-following decision-making modules and the commercially available Openpilot driving stack by CommaAI.
\end{abstract}


\end{frontmatter}

\section{Introduction}\label{sec:intro}
The car-to-car rear-end collision has been the most common crash type in the U.S. for decades. Various Advanced Driver Assist Systems (ADAS) have been developed and deployed to help mitigate the read-end collision risk, including crash-imminent braking (CIB), autonomous emergency braking (AEB), traffic jam assist (TJA), adaptive cruise control (ACC), and pedestrian crash avoidance mitigation (PCAM). In this paper, we are primarily interested in vehicle following ADAS, which cover a large portion of the currently available ADAS. We assume the Subject Vehicle (SV) is sufficiently well-performed in other operational modules, such as lane-keeping. This is a common assumption and is feasible to achieve in the practice of ADAS tests. We also emphasize that the proposed approach is applicable to evaluate other ADAS modules, such as the Lane-Keeping Assist System (LKAS), yet details are beyond the scope of this paper.

The safety evaluation of an ADAS-equipped SV in the car-following and rear-end collision avoidance regime seeks to characterize the SV's safety performance against stationary/moving vehicles in the front of the SV within the same lane, or along the SV's current trajectory. One common testing approach is to observe the SV's performance in the real-world or simulated naturalistic driving environment for a sufficiently long driving distance. One then observes or infers the collision rate estimate. This is formally known as the Monte-Carlo sampling, with other importance-sampling based variants from~\cite{zhao2017accelerated} that help improve the sampling efficiency. However, the required testing effort is still too significant to be widely applicable in practice. The naturalistic driving environment is not necessarily unchanged and may vary significantly from time to time. For those importance sampling based variants, the importance function estimate was developed with various heuristics, making it difficult to justify its accuracy. Also, as reported by~\cite{weng2021finite}, such a statistical inference method occurs in an implicitly defined operable domain with the tendency to over-estimate the risk. Finally, a simple scalar measure of risk is not necessarily sufficient to justify the complete safety performance of an SV.

The dominant approach adopted by most existing regulatory and standards follows the scenario-based test where the SV is deployed as a black-box system (uncontrollable and partially observable) in a testing case with the lead vehicle following a certain prescribed control policy. The common practice in this case presents a finite set of concrete scenarios and analyzes the testing outcome through an independent safety metric (i.e. the metric is computed independently from the test execution and data acquisition, and the testing data is presented as it stands). Some commonly observed concrete scenarios in the rear-end collision avoidance regime include the car-to-car lead vehicle braking in~\cite{forkenbrock2015nhtsa}, the suddenly revealed stationary vehicle (SRSV) and the lead vehicle lane change and brake (LVLCB) in~\cite{rao2019test}, also known as the frontal cut-in scenario, to name a few. The testing is mostly performed in a real-word proving grounds with a certain strikable target that emulates the motion and the appearance of a lead vehicle. Some also execute the test in a hardware-in-the-loop fashion such as the augmented scenes by~\cite{feng2020safety}. The results are then analyzed using an added metric, such as the observed collision rate, time-to-collision violation (TTCV) by~\cite{wishart2020driving}, and other surrogate measures summarized in~\cite{wang2021review}. Note that this is also the testing approach adopted by many regulatory standards such as the Europe NCAP by~\cite{ncapc2c2019}. However, as reported by~\cite{weng2021towards}, the set of concrete scenarios has very poor coverage of the SV's operational domain and is not of sufficient risk. The safety metrics are mostly biased and fail to arrive at a consensus agreement and make a fair comparison among various SVs as shown in~\cite{weng2021model}. The approach is also fundamentally problematic if the underlying system is stochastic which is a common phenomena in practice, and has been further enhanced as more learning-based methods are involved in perception and decision-making modules.

In this paper, we propose a scenario-sampling framework built on the Synchronous Pruning and Exploration (SPE) for safe set quantification in~\cite{weng2021towards} with various improvements dedicated to the car-following regime tests in practice. The basic idea of the proposed framework seeks to characterize the safe operational design domain (ODD) of the SV in the car-following regime through repeatedly sampling runs of scenarios in a guided manner. With a certain desired confidence level, one can then claim at \textit{what} states the SV is potentially safe and \textit{how} safe the SV is within the derived set of states. The proposed method is further demonstrated in Section~\ref{sec:case}, where it is shown capable of capturing various subtle safety properties and insights of widely adopted car-following models in both academic research as well as commercially available ADAS products in practice. The studied ADAS are more realistic and difficult to evaluate than some of the previous work by~\cite{fan2017d} and \cite{zhao2016accelerated}. To the best of knowledge, many of the obtained properties have never been captured by existing work in the field.

\textbf{Notation: } The set of real and positive real numbers are denoted by $\R$ and $\R_{>0}$ respectively. $\Z$ denotes the set of all positive integers and $\Z_N=\{1,\ldots,N\}$. $|\mathcal{X}|$ is the cardinality of the set $\mathcal{X}$.

\section{Preliminaries and Problem Formulation}\label{sec:prob}
Consider the general discrete-time system dynamics
\begin{equation}\label{eq:dyn}
    \s(t+1) = f(\s(t);\omega(t))
\end{equation}
with state $\s \in \Ss \subseteq \R^n$, uncertainties and disturbances $\omega \in \W \in \R^w$, for some $n, w \in \Z$. Let $\mathcal{C}\subset \Ss$ denote the set of failure states. Intuitively, for the system~\eqref{eq:dyn} to remain statistically safe, there should exist $\Ss^* \subset \Ss$, $\Ss^*\cap \mathcal{C} = \emptyset$ and all trajectories initialized in $\Ss^*$ remain inside $\Ss^*$ with high probability. The safety performance justification then seeks to characterize the set $\Ss^*$. In practice, $\Ss^*$ could be non-convex, non-unique, and of other complex structures, leading to various challenges for accurate characterization, statistically or deterministically. In this paper, we adopt the $\epsilon\delta$-almost safe set based methods from~\cite{weng2021formal}. Some important definitions and theorems are revisited in the following sub-section.

\subsection{$\epsilon\delta$-Almost Safe Set}
The following definition is adapted from~\cite{weng2021towards,weng2021formal}.
\begin{definition}\label{def:delta-cov-set}
    \textbf{($\delta$-Covering Set)} Give a compact set $\mathcal{X}\subset \R^n$ for some $n\in\Z$ and $\delta \in \R^n$. For any $\mathbf{x} \in \mathcal{X}$, let $\mathcal{N}_{\delta}(\mathbf{x})$ be the $\delta$-neighbourhood of $\mathbf{x}$, i.e., $\forall \mathbf{x}'\in \mathcal{N}_{\delta}(\mathbf{x}), |\mathbf{x}-\mathbf{x}'| \leq \boldsymbol\delta.$
    We claim that $\Phi_{\delta}^{\mathcal{X}}$ is a $\delta$-covering set of $\mathcal{X}$ if for some $k \in \Z$ and $\mathbf{x}_i\in\mathcal{X}, i=1,\ldots,k$, we have $
        \Phi_{\delta}^{\mathcal{X}}\!=\!\bigcup_{i\in\{1,\ldots,k\}} \mathcal{N}_{\delta}(\mathbf{x}_i) \supseteq \mathcal{X} \text{ and } \Phi_{\sigma}^{\mathcal{X}}\!=\!\{\mathbf{x}_i\}_{i\in\{1,\ldots,k\}} \subseteq \mathcal{X}.$
    Furthermore, $\Phi_{\sigma}^{\mathcal{X}}$ are centroids of $\Phi_{\delta}^{\mathcal{X}}$.
\end{definition}
Recall $\mathcal{C}$ is the set of failure states (e.g. collisions). The following definition formally characterizes the notion of the SV being ``almost" safe in a certain set.
\begin{definition}
    \textbf{($\epsilon\delta$-Almost Safe Set)} Given the system dynamics~\eqref{eq:dyn}, $\epsilon \in (0,1]$, $\delta\in\R^n$, $\Phi \subseteq \Ss$. The set $\Phi$ is $\epsilon\delta$-almost safe for the system~\eqref{eq:dyn} if there exists a $\delta$-covering set $\Phi_{\delta}$ of $\Phi$ with $\Phi_{\sigma}$ such that $\Phi_{\delta} \cap \mathcal{C} = \emptyset$ and
    \begin{equation}\label{eq:almost-cpis}
        \mathbb{P}\Big(\big\{\forall \s \in \Phi_{\sigma}, \forall \boldsymbol\omega \in \W : f(\s; \boldsymbol\omega) \not\in \Phi_{\delta}\big\}\Big)\leq \epsilon.
    \end{equation}
\end{definition}
It is immediate from the above definition that $\lim_{\boldsymbol\delta\rightarrow 0}\Phi_{\delta}^{\mathcal{X}} = \mathcal{X}$. Also note that as $\epsilon$ tends to zero, the $\epsilon\delta$-almost safe set becomes an absolutely safe $\delta$-covering set. To adapt the above definitions to the application of car-following regime safety analysis, we shall first characterize the car-following scenario in the form of~\eqref{eq:dyn}. 

\subsection{The Scenario-based Car-Following System}
In this paper, we consider the following system to formulate the interactive motion between a Subject Vehicle (SV) follower and a leading Principal Other Vehicle (POV) in the front sharing the same lane with the SV:
\begin{equation}\label{eq:sys_with_ctrl}
    \s(t+1) = f_s(\s(t), \uu(t); \omega_s(t)).
\end{equation}
The state $\s=[d, v_0, v_1] \in \Ss \subset \R_{\geq0}^3$, where $d \in [0, \infty)$ denotes the distance headway (simplified as headway or DHW in this paper) between the two vehicles, $v_0 \in [0,v_{\max}]$ and $v_1\in[0,v_{\max}]$ denote the longitudinal velocity of the SV follower and the lead POV, respectively. In practice, significantly large $d$ is not of safety concern, hence the upper bound of $d$ is often replaced with a sufficiently large value $d_{\max} \in \R_{>0}$. Other disturbances and uncertainties are denoted as $\omega_s\in \W_s$, which could involve environmental features (e.g. weather condition and road surface friction), infrastructure information (e.g. road curvature, road gradient, and speed limit), other kinematic and dynamic features (e.g. lateral offset between the vehicles and acceleration status of vehicles), other road users (e.g. pedestrian, cyclist, and other vehicles), planning parameters (e.g., free-traffic speed), and measurement error, to name a few. As also discussed by~\cite{weng2021finite}, the state $\s$ and some of the uncertainties $\omega_s$ may be interchangeable depending on the particular feature's observability and how important it is in determining safety related properties. For example, \cite{ncapc2c2019} consider the lateral offset between vehicles as an important feature that affects the performance of SV, leading to an extra dimension added to the state $\s$. The action $\uu \in \U \subset \R$ represents the control input of the lead POV, such as the desired velocity and the commanded acceleration. Note that the SV is the test subject in the testing content, thus it is an uncontrollable and (partially) observable black-box system (see Remark 3 in~\cite{weng2021towards}). Furthermore, the action $\uu$ is typically determined by a certain feedback control policy
\begin{equation}\label{eq:ctrl}
    \uu = \pi(\s, \omega_s; \omega_u),
\end{equation}
with $\s$, $\omega_s$ the same with what we have defined above, and the uncertainties $\omega_u \in \W_u$. Intuitively, the policy $\pi$ describes the lead POV driving behavior. In the scenario-based safety evaluation regime, the testing policy is a given function. As a result, composing~\eqref{eq:sys_with_ctrl} with \eqref{eq:ctrl} we have the exact system dynamics of~\eqref{eq:dyn} with $n=3$. The disturbances and uncertainties $\omega\in \W$ is jointly affected by $\s, \omega_s$ in~\eqref{eq:sys_with_ctrl} and $\omega_u$ in~\eqref{eq:ctrl}. In practice, the scenario system may not necessarily exhibit the Markov Decision Process (MDP) nature induced by~\eqref{eq:dyn} as the next-step state may be dependent upon not only the current state, but also a series of historic observations. One can extend the state space to involve those observations, yet the state space complexity will also increase significantly. In the particular car-following domain studied by this paper, we argue that the capability of SV taking advantage of historical information, if applicable, would only make a better safety performance. As a result, the safety property obtained from system~\eqref{eq:dyn} still remains as the worst-case justification.

A run of a test scenario, $\mathcal{RS}(\s_0, K)$ ($K \in \Z, K\geq 2$), thus starts from a certain state initialization $\s_0\in \Ss$, consecutively collects a set of states admitting the system dynamics~\eqref{eq:dyn}, and terminates either when encountering a failure event (e.g., collision) or the $K$-th step of observation is reached. If $f$ is explicitly known or approximately characterized, one can execute the test scenario and collect data through computer simulations. On the other hand, the scenario-based test can also be performed in real-world testing proving ground with $f$ implicitly induced.

The standard scenario-based safety evaluation methods (e.g. NCAP~\cite{ncapc2c2019} and NHTSA guidelines in~\cite{forkenbrock2015nhtsa,rao2019test}) specify the $\s_0$ based on expert-knowledge and real-world crash database. The test policy $\pi(\cdot)$ is typically presented as a deterministic function with constant deceleration magnitude (e.g. $\pi(\s) = -6 \mathrm{ m/s^2}, \forall \s \in \Ss$ in some of the car-to-car AEB cases). In this paper, we adopt a similar design of $\pi(\cdot)$ used by the above mentioned standardized tests (i.e., the lead POV executes the braking maneuver at a constant deceleration rate). This evaluates the SV's safety performance in a more adversarial environment than the naturalistic driving environment. We also emphasize that the proposed method does not rely on a particular testing policy, and will generalize easily to other testing policies, such as those emulating naturalistic driving behaviors in~\cite{zhao2016accelerated}.
\subsection{The Almost Safe Set Quantification Problem}
Let a scenario-sampling algorithm consecutively sample runs of scenarios on $\Ss$ following the system dynamics~\eqref{eq:dyn}. We are now ready to present the car-following safe set quantification problem as follows.
\begin{problem}\label{prob}
    Given $\delta\in\R^n$, $\epsilon\in(0,1], \beta(0,1]$, a testing policy $\pi(\cdot)$ in the form of~\eqref{eq:ctrl}, and the corresponding car-following scenario system in the form of~\eqref{eq:dyn}. Let $\Ss_0 \subseteq \Ss$ be the sup-set of all safe sub-sets in $\Ss$. The \textit{car-following safe set quantification problem} seeks to find a scenario-sampling algorithm $\mathcal{ALG}: \Ss\times(0,1]\times(0,1]\times\R^n \rightarrow \Ss$,
    such that with confidence level at least $1-\beta$, $\mathcal{ALG}(\Ss_0,\epsilon,\delta,\beta)$ is an $\epsilon\delta$-almost safe set for~\eqref{eq:dyn}.
\end{problem}
The previous work by~\cite{weng2021towards} has already presented various algorithms that provably solve the above problem with a primary focus on completeness and asymptotic optimality properties. Such properties occur as the number of samples tends to infinity which leads to a significant amount of samples required in practice. In this paper, we propose a modified version of the Synchronous Pruning and Exploration for safe set quantification by~\cite{weng2021towards} with a specific focus on the car-following regime. This leads to a theoretically sound and practically feasible safe set quantification solution as we shall see in the next two sections.

We conclude this section by addressing the following assumption and justifying its practical feasibility.
\begin{assumption}
    Given the state space $\Ss$, the set of failure states $\mathcal{C}$, and the system~\eqref{eq:dyn}, we assume that the run of scenario can be initialized from any $\s \in \Ss\setminus \mathcal{C}$.
\end{assumption}
In practice, if one can control the engagement of the subject ADAS sufficiently accurately, the above assumption is naturally feasible, such as the test protocol by~\cite{ncapc2c2019}. On the other hand, if the ADAS is expected to engage before triggering the test, the accurate initialization becomes more difficult at some states. In this case, the above assumption is easy to achieve mostly at the control equilibrium sub-set of $\Ss$. For example, $v_0=v_1$ for some $d \in \R_{>0}$, which denotes the steady-state car-following scene. This is also the initialization condition adopted by~\cite{forkenbrock2015nhtsa}. Some non-control equilibrium states can be initialized through customized scenes. For example, in the LVLCB test from the NHTSA report by~\cite{rao2019test}, the lead-vehicle on the side lane can choose to perform a lane change at any speed with any headway, which has the potential to initialize some non-control equilibrium states such as when $v_0 \gg v_1$. Note that even with the above techniques, some states are still difficult to initialize, such as $v0\gg v1, d=0$. However, those difficult-to-achieve initialization states are typically of obvious high-risk, thus they may not need to be tested anyway, as we shall see in Section~\ref{sec:case}. 

\section{Main Method}\label{sec:method}
To solve Problem~\ref{prob}, the overall algorithm follows a two-step procedure. First, one continuously constructs a candidate set as more runs of scenarios are collected through scenario sampling. Second, as the constructed set becomes close to the actual almost safe set, one should observe a sufficiently large number of runs of scenarios that start from and remain inside the candidate set. For the second step, the sampling sufficiency is justified by the following theorem.

\begin{theorem}\label{thm:validation}
    \textbf{($\epsilon\delta$-Almost Safe Set Validation)} Given the system dynamics~\eqref{eq:dyn}, $\epsilon \in (0,1]$, $\beta \in (0,1]$, $\delta\in\R^n$, $\Phi \subseteq \Ss$, and the corresponding $\delta$-covering set $\Phi_{\delta}$ with centroids $\Phi_{\sigma}$ defined by Definition~\ref{def:delta-cov-set}. Consider $N$ runs of scenarios, $\{\mathcal{RS}_i(\s_0, K)\}_{i=1,\ldots,N}$ ($K\in\Z, K\geq2$), with the state initialization of each run being i.i.d. w.r.t. the underlying distribution on $\Phi_{\sigma}$. The set $\Phi$ is the $\epsilon\delta$-almost safe set for~\eqref{eq:dyn} with confidence level at least $1-\beta$ if $\bigcup_{i=1}^N \mathcal{RS}_i(\s_0, K) \subseteq \Phi_{\delta} \cap \mathcal{C}=\emptyset$ and $N \geq \frac{\ln{\beta}}{\ln{(1-\epsilon)}}.$
\end{theorem}
That is, under the given conditions, if one consecutively observes $\frac{\ln{\beta}}{\ln{(1-\epsilon)}}$ runs of scenarios remaining inside $\Phi_{\delta}$, one then have the confidence level at least $1-\beta$ to claim that the probability for any trajectory starting from $\Phi_{\sigma}$ to leave $\Phi_{\delta}$ is less than $\epsilon$, i.e., the SV is $\epsilon\delta$-almost safe in the set $\Phi$. One can refer to~\cite{weng2021towards} for the proof of Theorem~\ref{thm:validation}.

The proposed algorithm to solve Problem~\ref{prob} is presented as Algorithm~\ref{alg:qnt} taking advantage of the Theorem~\ref{thm:validation}. Note that \texttt{pop}, \texttt{reachable}, \texttt{nearest}, \texttt{remove}, and \texttt{append} are all notional functions. $\mathcal{X}.$\texttt{pop}() returns a point $\x\in\mathcal{X}$ and removes it from the set. \texttt{reachable}($\s, G$) returns all vertices on the graph $G$ that connects, directly and indirectly, to the point $\s$ through a depth-first-search routine (see~\cite{sdq_tools}). $\mathcal{X}$.\texttt{nearest}($\x$) returns the nearest point to $\x$ in $\mathcal{X}$ in terms of $\ell_2$-norm distance. The commands \texttt{remove} and \texttt{append} simply remove a point from or add a point to the given set, respectively.

Overall, Algorithm~\ref{alg:qnt} consists of four major steps. The initialization step (line 2) configures two graphs, $G_{\sigma}$ and $G_u$, that are intended to contain potentially safe and observed unsafe states and transitions, respectively, through scenario-sampling. The sampling step (line 4-7) takes a i.i.d. sample by Theorem~\ref{thm:validation} if the prioritized replay buffer $\mathcal{B}$ is empty. Otherwise, i.e. when some unsafe states have been observed and added to $\mathcal{B}$ at line 12, it  prioritizes sampling points in $\Phi_{\sigma}$ that are close to the points in $\mathcal{B}$ as they are intuitively of higher-risk. Such a sampling heuristic will not jeopardize the claimed property in Theorem~\ref{thm:validation} for set validation, as $\mathcal{B}$ will be empty eventually, but will accelerate the convergence to a sufficiently almost safe set as unsafe points are removed more frequently. The third important stage happens at line 10-19. When a sampled run of a scenario is observed to converge to $\mathcal{C}$, any reachable states to the points in the collected run are removed from $\Phi_{\sigma}$. On the other hand (line 21-32), one either adds an uncovered point to the covering set (line 23-25) or consecutively observes $N$ runs of scenarios that remain inside $\Phi_{\delta}$ to claim the $\epsilon\delta$-almost safe property.

The proposed algorithm differs from the SPE for safe set quantification in~\cite{weng2021towards} in two main ways, the use of prioritized sampling with a replay buffer and the removed stage of $\epsilon\delta$ decay. The prioritized sampling with a replay buffer is a heuristic approach that improves the convergence rate to a potentially almost safe set. The fixed choice of $\delta$ and $\epsilon$ compromises the probabilistic completeness of the algorithm in return for  practical feasibility with improved sampling efficiency (as we shall also see empirically in Section~\ref{sec:case}). One can always re-obtain the completeness and optimality properties, or at least achieve an appropriate level of compromisation, by configuring $\delta$ and $\epsilon$ to be arbitrarily close to zero, yet the number of required samples might also increase dramatically.

\begin{algorithm}[H]
    \begin{algorithmic}[1]
    \State {\bf Input:} Initial set $\Ss_0\subseteq\Ss$, collision set $\mathcal{C}$, $\epsilon\in(0,1]$, $\beta\in(0,1]$, trajectory horizon $K$.
    \State {\bf Initialize: } The $\delta$-covering set of $\Ss_0$, $\Phi_{\delta}$, and centroids $\Phi_{\sigma}$ by Definition~\ref{def:delta-cov-set}, the state graph $G_{\sigma} = (\Phi_{\sigma}, E_{\sigma}), E_{\sigma} =\emptyset
    \subset \Ss^2$, the unsafe state graph $G_u=(\mathcal{D}_u, E_u), \mathcal{D}_u=\emptyset\subset \Ss, E_u = \emptyset \subset \Ss^2$, prioritized replay buffer $\mathcal{B}=\emptyset$, N=0.
    \State{{\bf While} $N<\frac{\ln{\beta}}{\ln{(1-\epsilon)}}$:}
    \State{\ \ \ \ {\bf If} $\mathcal{B}=\emptyset$}
    \State{\ \ \ \ \ \ \ \ $\s_0 \sim P(\Phi_{\sigma})$}
    \State{\ \ \ \ {\bf Else}}
    \State{\ \ \ \ \ \ \ \ $\s_b = \mathcal{B}$.\texttt{pop}(), $\s_0 = \Phi_{\sigma}.$\texttt{nearest}($\s_b$)}
    \State{\ \ \ \ {\bf End If}}
    \State{\ \ \ \ Get $\mathcal{T}=\mathcal{RS}(\s_0, K)$}
    \State{\ \ \ \ {\bf If} $\mathcal{T}\cap\mathcal{C} \neq \emptyset$}
    \State{\ \ \ \ \ \ \ \ {\bf For $i$ in $\Z_{|\mathcal{T}|-1}$} {\bf do}}
    \State{\ \ \ \ \ \ \ \ \ \ \ \ $\mathcal{B}$.\texttt{append}($\mathcal{T}[i]$)}
    \State{\ \ \ \ \ \ \ \ \ \ \ \ {\bf For} $\s$ in \texttt{Reachable}($\mathcal{T}[i],G_{\sigma})$ {\bf do}}
    \State{\ \ \ \ \ \ \ \ \ \ \ \ \ \ \ \ $\Phi_{\sigma}$.\texttt{remove}($\s$)}
    \State{\ \ \ \ \ \ \ \ \ \ \ \ $E_u$.\texttt{append}(($\mathcal{T}[i]$, $\mathcal{T}[i+1]$))}
    \State{\ \ \ \ \ \ \ \ \ \ \ \ {\bf End For}}
    \State{\ \ \ \ \ \ \ \ $\mathcal{B}$.\texttt{append}($\mathcal{T}[i+1]$)}
    \State{\ \ \ \ \ \ \ \ {\bf End For}}
    \State{\ \ \ \ \ \ \ \ $N=0$}
    \State{\ \ \ \ {\bf Else}}
    \State{\ \ \ \ \ \ \ \ $\bar{\s}=\s_0, N_s=|\Phi_{\sigma}|$}
    \State{\ \ \ \ \ \ \ \ {\bf For $i$ in $\{2,\ldots,|\mathcal{T}|\}$} {\bf do}}
    \State{\ \ \ \ \ \ \ \ \ \ \ \ {\bf If} $\mathcal{T}[i] \notin \Phi_{\delta}$}
    \State{\ \ \ \ \ \ \ \ \ \ \ \ \ \ \ \ $E_{\sigma}$.\texttt{append}(($\bar{\s}$, $\mathcal{T}[i]$))}
    \State{\ \ \ \ \ \ \ \ \ \ \ \ \ \ \ \ $\bar{\s}=\mathcal{T}[i]$}
    \State{\ \ \ \ \ \ \ \ \ \ \ \ {\bf End If}}
    \State{\ \ \ \ \ \ \ \ {\bf End For}}
    \State{\ \ \ \ \ \ \ \ {\bf If} $N_s=|\Phi_{\sigma}|$ and $\mathcal{B}=\emptyset$}
    \State{\ \ \ \ \ \ \ \ \ \ \ \ $N += 1$}
    \State{\ \ \ \ \ \ \ \ {\bf Else}}
    \State{\ \ \ \ \ \ \ \ \ \ \ \ $N = 0$}
    \State{\ \ \ \ \ \ \ \ {\bf End If}}
    \State{\ \ \ \ {\bf End If}}
    \State {{\bf Output:} $\Phi_{\delta}$}
    \end{algorithmic}
    \caption{Car-following Safe Set Quantification} \label{alg:qnt}
\end{algorithm}

\section{Case Studies}\label{sec:case}
To demonstrate the performance of the proposed Algorithm~\ref{alg:qnt}, we start with examples of safety evaluations of deterministic decision-making systems where the perception and the control modules are both sufficiently accurate. We then move to an end-to-end case study taking the CommaAI's Openpilot by~\cite{openpiotgithub} as an example which involves a neural-network based perception module, camera-radar sensor fusion, model-based decision-making, and control modules. The source code for Algorithm~\ref{alg:qnt} in Python can be found at~\cite{sdq_tools}.

\subsection{Decision-Making Safety Evaluation}\label{sec:dm_case}
We consider two classes of decision making systems in this section. The first is a combination of ACC and AEB (ACC-AEB) first introduced by~\cite{zhao2016accelerated}. When the perceived time-to-collision value is greater than a pre-determined threshold, the ACC module is engaged as a discrete Proportional-Integral (PI) controller to achieve a desired time headway. Otherwise, the AEB module extracted from a 2011 Volvo V60 is active. The ACC-AEB module takes the same hyper-parameters and configuration as~\cite{zhao2016accelerated}, having a maximum braking capability of $-10 \mathrm{ m/s^2}$ subject to a deceleration change rate limit of $-16 \mathrm{ m/s^3}$. The second decision-making module studied by this section is the Intelligent Driving Model (IDM) in~\cite{treiber2013traffic}, which is a widely adopted car-following model in the field. Note that we have created three IDM variants based on the maximum brake control capability. In particular, we have the normal-brake IDM (N\_IDM) with $-5 \mathrm{ m/s^2}$, the mild-brake IDM (M\_IDM) with $-3 \mathrm{ m/s^2}$, and the hard-brake IDM (H\_IDM) with $-7 \mathrm{ m/s^2}$. Other IDM parameters include the minimum safe distance (2 m), maximum acceleration (0.73$\mathrm{ m/s^2}$), comfortable deceleration (1.67$\mathrm{ m/s^2}$), safe time headway (2 s), exponent of acceleration (4), and vehicle length (4 m). Unless mentioned otherwise, we consider the state space $\Ss$ with the headway $d \in [0,100]$ m, SV speed $v_0 \in [0,30]\text{ m/s}$, and lead POV speed $v_1 \in [0,30]\text{ m/s}$. Note that the collected run of a scenario might leave $\Ss$ with a large headway value that is greater than the given upper bound (100 m), in which case, one shall either truncate the trajectory or clip the headway value at the given upper bound before proceeding to line 10 of Algorithm~\ref{alg:qnt}. The simulation of each run of scenario operates at $10$ Hz with $K=300$. The testing policy admits the form of $\pi(\s)=-5 \mathrm{ m/s^2}, \forall \s \in \Ss$. We also assume the free-traffic speed to be $30 \text{ m/s}$.

\begin{figure}[!t]
\centering
\includegraphics[trim={3cm 0cm 4cm 1cm},clip,width=0.46\textwidth]{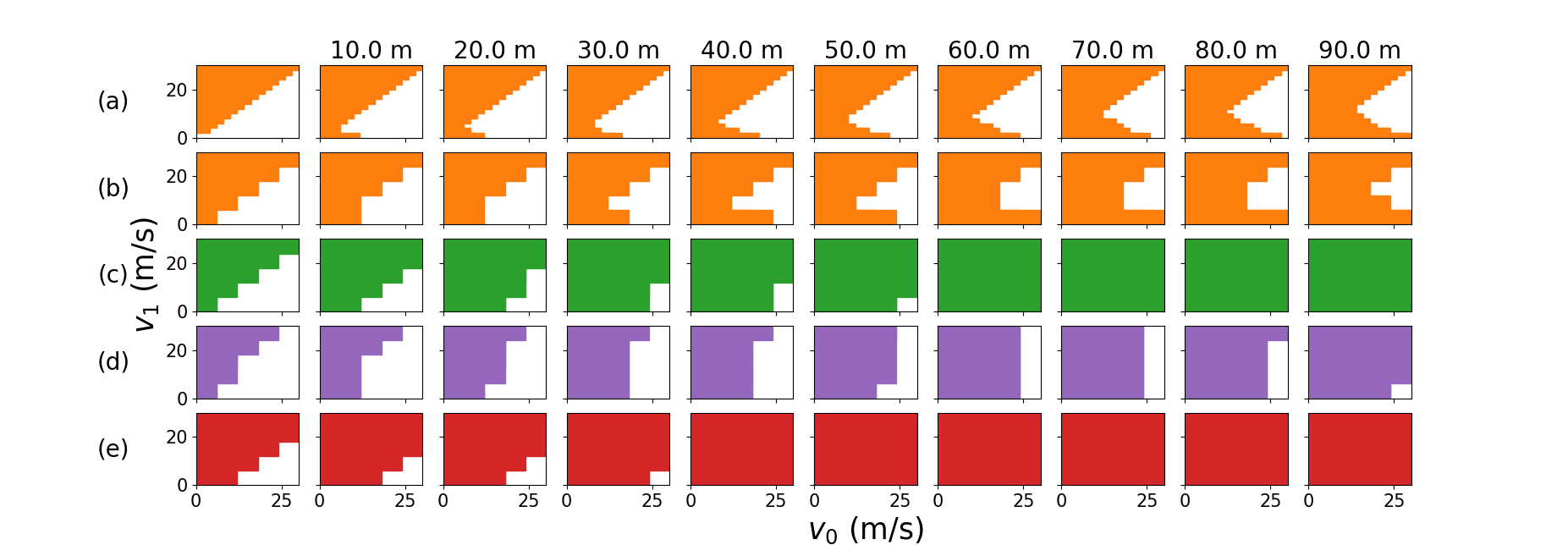}
\vspace{-4mm}
\caption{\footnotesize{Some $\epsilon\delta$-almost safe sets obtained for the car-following case study with various decision-making modules ($\epsilon=0.01, \beta=0.001$): (a) ACC-AEB with $\delta=[10,2,2]$, (b) ACC-AEB with $\delta=[10,6,6]$, (c) N\_IDM with $\delta=[10,6,6]$, (d) M\_IDM with $\delta=[10,6,6]$, (e) H\_IDM with $\delta=[10,6,6]$.}}
\label{fig:lead_follow}
\vspace{-2mm}
\end{figure}

\begin{table}
\begin{center}
\caption{\footnotesize{The safety evaluation results for various decision-making modules in the car-following case ($\beta=0.001, \delta=[10,6,6]$) and Openpilot presented in Section~\ref{sec:e2e_case} ($\beta=0.001, \delta=[3,3,3]$).}}\label{tb:lead_follow}
\resizebox{0.45\textwidth}{!}{%
\begin{tabular}{cccccc}
SV & $\Ss_0$ & $\epsilon$ & scenario runs & collision runs& IoU \\\hline
ACC-AEB & $\Ss$ & $0.1$ & $867.5\pm 281.2$ & $268.3\pm 34.5$  & $0.915$\\
\       & $\Ss$& $0.01$ & $1912.6\pm 146.4$ & $185.4\pm 1.4$  & $1.000$\\\hline
H\_IDM  & $\Ss$& $0.1$ & $194.2\pm 14.6$ & $40.7\pm 3.2$  & $0.965$\\
\       & $\Ss$& $0.01$ & $1376.0\pm 182.1$ & $49.0\pm 0.0$  & $1.000$\\\hline
N\_IDM  & $\Ss$& $0.1$ & $368.5\pm 95.0$ & $69.6\pm 4.4$  & $0.952$\\
\       & $\Ss$& $0.01$ & $1628.8\pm 266.3$ & $74.6\pm 0.8$  & $0.998$\\
\       & Fig~\ref{fig:lead_follow}e& $0.01$ & $1578.6\pm 220.1$ & $26\pm 0.0$  & $1.000$\\\hline
M\_IDM  & $\Ss$& $0.1$ & $830.9\pm 88.3$ & $155.6\pm 3.7$  & $0.956$\\
\       & $\Ss$& $0.01$ & $1892.6\pm 237.5$ & $161.0\pm 0.0$  & $1.000$\\
\       & Fig~\ref{fig:lead_follow}e& $0.01$ & $1731.4\pm 125.5$ & $112.0\pm 0.0$  & $1.000$\\\hline
Openpilot & $\Ss$ & $0.1$ & $704.2\pm 54.3$ & $141.8\pm 3.4$  & $0.897$ \\ \hline 
\end{tabular}}
\end{center}
\end{table}

We execute Algorithm~\ref{alg:qnt} for 10 times with 10 different random seeds. The set of 10 seeds remains the same among different SVs. Some of the obtained almost safe sets for $\epsilon=0.01, \beta=0.001$ are illustrated in Fig~\ref{fig:lead_follow} for the same seed. The three-dimensional safe set is illustrated with a series of subplots on the $(v_0, v_1)$ domain, each representing a subspace slicing of a certain headway value. Intuitively, the size of the safe set increases as the lead-POV becomes further away, since the state is of lower-risk as the lead-POV operates at a higher speed than the SV follower. This is mostly correct if one observes the IDM cases where M\_IDM has the smallest almost safe set and H\_IDM has the largest almost safe set, which aligns with the underlying configurations of M\_IDM having the lowest braking capability and H\_IDM having the strongest braking capability among all tested IDMs.
\begin{figure}[!t]
    \centering
    \includegraphics[trim={0cm 0cm 0cm 1cm},clip,width=.35\textwidth]{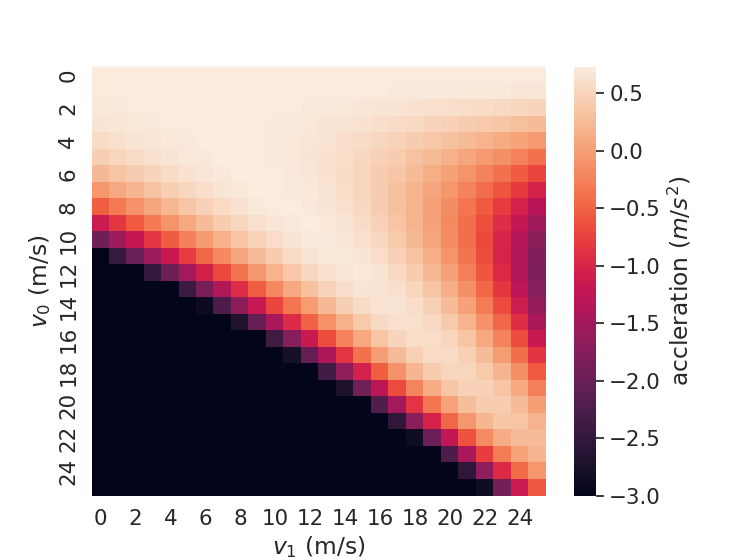}
    \vspace{-2mm}
    \caption{\footnotesize{M\_IDM's commanded acceleration inputs for a group of $(v_0,v_1)$ pairs at 40-meter headway in the car-following scenario.}}
    \label{fig:midm_40}
\end{figure}

However, for most of the subplots in the ACC-AEB case, especially those with large headway values, one exhibits a non-convex almost safe set with a white notch, which indicates some unsafe states even when the headway is sufficiently large. This is mainly due to the ACC design nature where one tends to reach the free-traffic speed aggressively when the headway value is high, $v_0$ thus increases, ending up in a certain unsafe state. For a similar cause, ACC-AEB also fails all of the CCRb and CCRm tests in Fig~\ref{fig:ncap}. As a result, if one considers the free-traffic speed as an observable state and expands the $\Ss$ to be of dimension four, the corresponding almost safe set will also change w.r.t. the desired velocity. A detailed analysis regarding this variant, and possibly other variants considering different added features, are of future interest. 

Returning to the notch observation, why isn't a similar shape showing up on any of the IDM variants in Fig~\ref{fig:lead_follow}? This is because the IDM is primarily a car-following model and may not necessarily exhibit expected behaviors outside the normal car-following work domain. For example, Fig~\ref{fig:midm_40} illustrates the M\_IDM's acceleration outputs for a group of $(v_0, v_1)$ pairs with 40-meter headway. Note that at $v_0=12$ m/s, $v_1=25$ m/s, the M\_IDM decides to execute maximum brake maneuver, rather than to accelerate to track the desired speed. This leads to a utility performance degradation in terms of velocity tracking, but on the other hand, improves the safety performance against potential rear-end collisions. Fundamentally speaking, the observed phenomena is caused by a squared term associated with the $(v_0-v_1)$ term in the IDM formulation, the details are beyond the scope of this paper. 

Moreover, comparing Fig~\ref{fig:lead_follow}(a) and Fig~\ref{fig:lead_follow}(d), the ACC-AEB has a relatively larger safe set than M\_IDM when the headway value is small. As the headway value increases, the safe set of M\_IDM enlarges significantly and eventually out-performs ACC-AEB in terms of the safe set size. That is, the notion of ``one vehicle being safer than the other" can be problematic as it is essentially a multi-dimensional comparison. A similar point was also made by~\cite{weng2021finite} through observing real-world car-following performance in the naturalistic driving environment. Such a subtle safety characterization is difficult to obtain by existing concrete scenario-based testing strategies such as the NCAP AEB testing shown in Fig~\ref{fig:ncap}.

More detailed results regarding this case are listed in Table~\ref{tb:lead_follow}. The ``IoU" denotes the intersection-over-union ratio of all obtained safe sets from different seeds w.r.t. the same SV. It is clear that the higher the IoU value is, the more similar the obtained sets are among different seeds. Considering the studied decision-making modules in this section are both deterministic, the IoU value should converge to one for sufficiently small $\epsilon$ and $\beta$. This has been validated empirically by row 2, 4, 7, 9, and 10 in Table~\ref{tb:lead_follow}. We emphasize that even for the cases with IoU values less than one, the results are not wrong, as the $\epsilon\delta$-almost safe set is simply not unique for the studied system. Also, note that if the set initialisation is not $\Ss_0$ but is another set that is closer to the final almost safe set, one should expect a smaller number of runs of scenarios and, more importantly, a smaller number of runs of scenarios with collisions, to converge to the desired outcome (e.g. comparing row 6 with row 7, and comparing row 9 with row 10 in Table~\ref{tb:lead_follow}). 

Overall, the total number of runs of scenarios varies w.r.t. the SV, the selected hyper-parameters (e.g. $\epsilon, \beta$) and the random seed, but remains below 2000 (i.e., less than 17-hour (2000 runs of scenarios with at most 30 seconds for each run) of actual scenario-running time excluding the testing preparation and scenario restoration time). This is a slightly higher testing burden than the existing standards for the car-following regime but should still be considerred feasible in practice. One can improve the efficiency in computer simulations by executing multiple testing scenarios in parallel. Moreover, the testing effort may be further reduced for a smaller $K$, and the exploration regarding this direction is of future interest. More importantly, among the methods that are capable of providing similar theoretical guarantees, the proposed solution appears to be the most practical and is capable of capturing the subtle differences among various SVs. For comparison, the importance sampling and Monte-Carlo sampling based methods reported by \cite{zhao2017accelerated} require hundreds of millions of test runs in simulation for safety evaluation with car-following maneuvers and only generate a risk estimate.

\subsection{End-to-End Safety Evaluation}\label{sec:e2e_case}
\begin{figure}[!t]
    \centering
    \includegraphics[trim={3cm 0cm 4cm 1cm},clip,width=.45\textwidth]{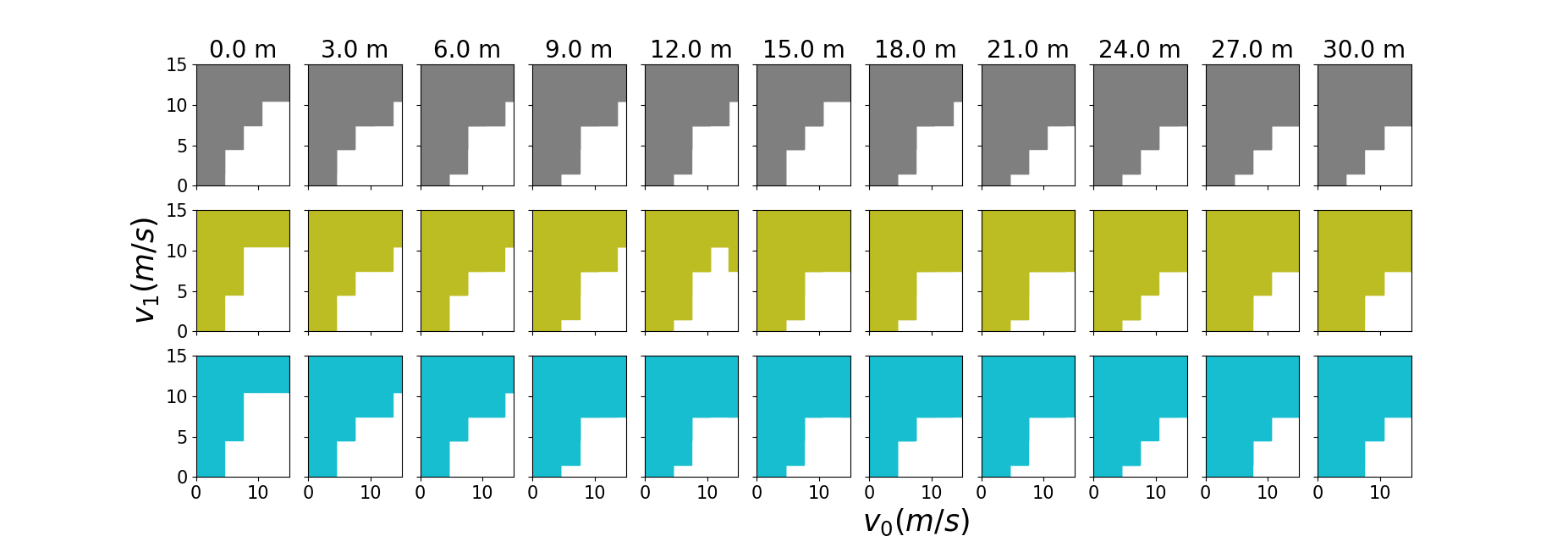}
    \vspace{-3mm}
    \caption{\footnotesize{Some $\epsilon\delta$-almost safe sets obtained for the car-following case study with Openpilot for three different random seeds ($\epsilon=0.1, \beta=0.001$) }}
    \label{fig:op_leadfollow}
    \vspace{-7mm}
\end{figure}

For an end-to-end case study, we evaluate the CommaAI Openpilot's safety performance in the car-following regime through simulation using the Carla simulator. To run the Openpilot in Carla, we use the Openpilot-Carla bridge provided by CommaAI as a foundation with added clustered radar results for radar-camera fusion to enable the ACC in Openpilot. The radar points clustering configuration is identical to the work by~\cite{zhong2021detecting}. The detailed implementation can be found at~\cite{openpilot_carla}. The state space $\Ss$ takes the configuration $d \in [0,30]$ m, $v_0 \in [0,15]\text{ m/s}$, and $v_1 \in [0,15]\text{ m/s}$. The simulation of each run of scenario operates at 100 Hz with $K=500$. The free-traffic speed is 11.176 m/s (25 mph) if $v_0(0) < 11.176$ and $v_0(0)$ otherwise, which is the default configuration of Openpilot. 

\begin{figure}
\centering
\subfloat[ACC-AEB.\label{fig:lo_acc_aeb}]{%
  \includegraphics[trim={1cm 0cm 2cm 2cm},clip,width=0.2\textwidth]{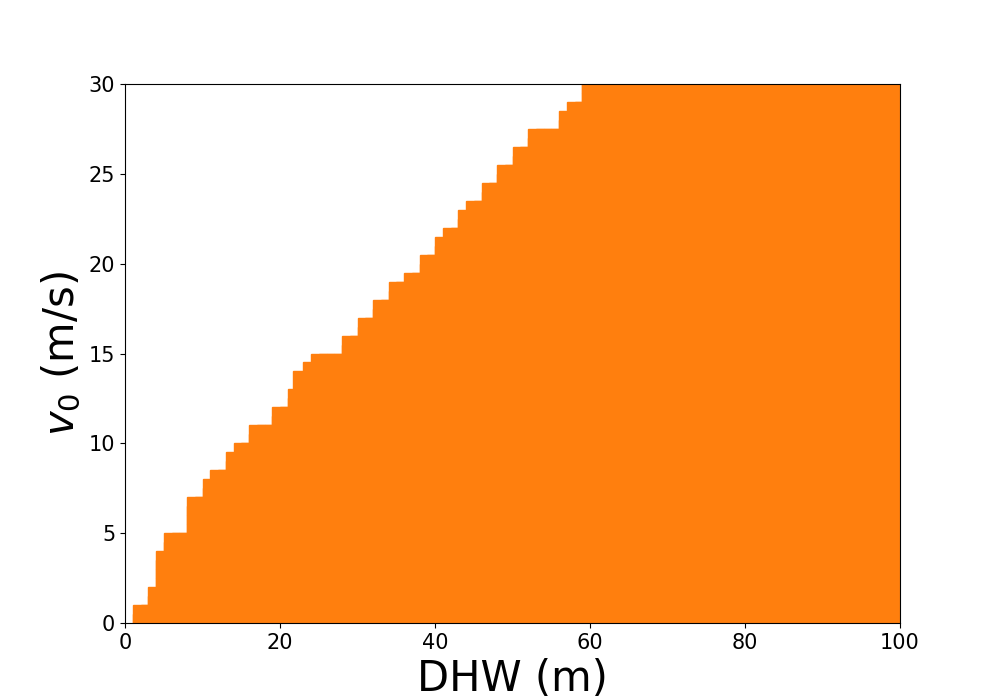}
}
\subfloat[Openpilot.\label{fig:lo_openpilot}]{%
  \includegraphics[trim={0cm 0cm 2cm 1cm},clip,width=.23\textwidth]{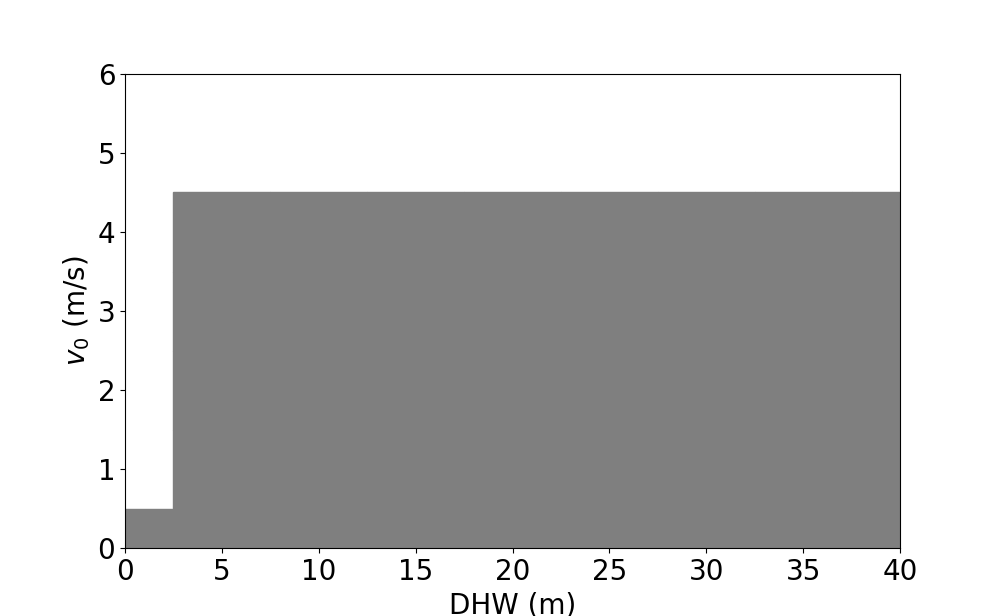}
}
\vspace{-3mm}
\caption{\footnotesize{The $\epsilon\delta$-almost safe sets ($\epsilon=0.01, \beta=0.001$) obtained for ACC-AEB and Openpilot in a lead-obstacle scene where the lead-POV remains stationary for all time.}}
\label{fig:lead_obstacle}
\end{figure}

\begin{figure}
    \centering
    \includegraphics[trim={1cm 0cm 6cm 0cm},clip,width=.45\textwidth]{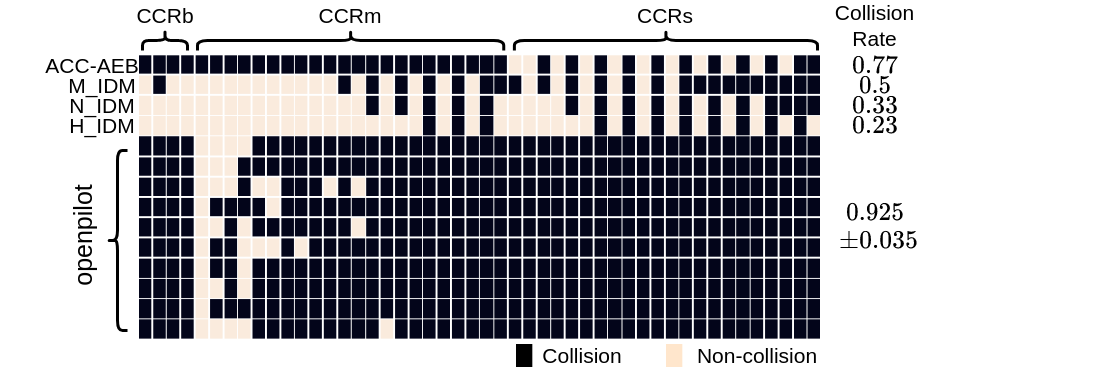}
    \vspace{-4mm}
    \caption{\footnotesize{The testing outcomes of all studied SVs in Section~\ref{sec:case} with the standard NCAP car-to-car AEB testing procedure discussed in~\cite{ncapc2c2019}. The procedure specifies 48 different scenario configurations from three categories including the Car-to-Car Rear stationary (CCRs), Car-to-Car Rear moving (CCRm), and Car-to-Car Rear braking (CCRb), where the lower-case letter after CCR induces the lead-POV's driving behavior (staying stationary, moving at a constant velocity, or braking to stop). Each deterministic decision-making module is only tested once. The Openpilot enabled SV is tested with the same set of 48 scenarios for 10 times. The detailed parameters related to the order of all testing cases can be found in ``ncap\_bridge.py" at~\cite{openpilot_carla}}.}
    \label{fig:ncap}
\end{figure}

Note that Openpilot is not designed for emergency collision avoidance as suggested by CommaAI at~\cite{openpiotgithub}. It is primarily a car-following model. As a result, an adversarial testing policy, such as the one adopted for the decision-making case, could lead to a very limited safe set. For example, as shown in Fig~\ref{fig:lead_obstacle}, if the lead vehicle remains stationary (similar to the CCRs case by~\cite{ncapc2c2019} and also included in Fig~\ref{fig:ncap}), the Openpilot SV almost fails to avoid any rear-end collisions if $v_0\geq4.5\mathrm{ m/s}$. The Openpilot's almost safe set is also significantly smaller than a regular almost safe set in cases such as the one shown for ACC-AEB in Fig~\ref{fig:lo_acc_aeb}. In this section, we admit the testing policy as $\pi(s)=0$ $ \mathrm{m/s^2}$, which emulates the steady-state car-following situation. 

\begin{figure}
    \centering
    \includegraphics[trim={0cm 0cm 0cm 0cm},clip,width=.4\textwidth]{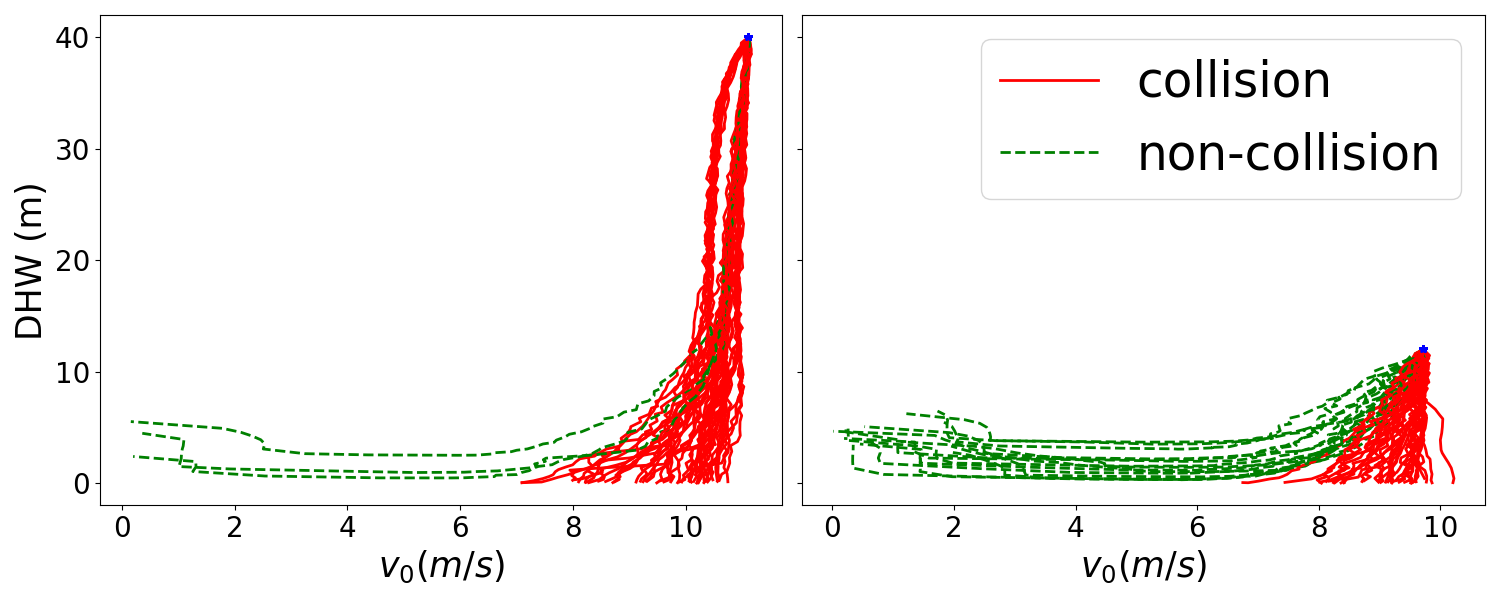}
    \vspace{-4mm}
    \caption{\footnotesize{The trajectories on the $(v_0, d)$ domain of Openpilot tested in two standard NCAP Car-to-Car Rear moving scenarios. For both scenarios, the lead POV remains at $20$ km/h ($5.56$ m/s). All other parameters and environmental configurations remain identical among all test runs for the same initialization condition. Within each subplot, Openpilot is enabled at the illustrated initialization state and both vehicles, unless specified otherwise by the testing procedure, maintain at the steady-state stage with zero acceleration.}}
    \label{fig:op_ncap}
    \vspace{-1mm}
\end{figure}

We execute Algorithm~\ref{alg:qnt} for 5 times with 5 different random seeds. Some of the obtained almost safe sets for $\epsilon=0.1, \beta=0.001$ with three different seeds are illustrated in Fig~\ref{fig:op_leadfollow}. Other statistical properties are summarized in the last row of Table~\ref{tb:lead_follow}. Note that the IoU rate in Table~\ref{tb:lead_follow} is slightly smaller than the presented cases in Section~\ref{sec:dm_case}. This is mainly due to the fact that Openpilot is fundamentally stochastic, as also illustrated by Fig~\ref{fig:op_ncap} and Fig~\ref{fig:ncap} where, starting from the same $\s_0$, the Openpilot enabled SV is shown capable of generating both safe and collision outcomes. As a result, the almost safe set for Openpilot in the studied domain is fundamentally non-unique, making it a particularly challenging case for many existing scenario-based techniques and surrogate safety metrics. As for the proposed method, the obtained safe set aligns with the claimed operational domain by CommaAI. The SV remains safe with high probability when $v_0 \geq v_1$ regardless of the following distance. The size of the almost safe set also increases as the headway value becomes larger.



\section{conclusion}\label{sec:conc}
In this paper, we have presented a theoretically sound and sampling efficient scenario-sampling framework for the safety performance evaluation of various car-following and rear-end collision avoidance systems. The performance of the proposed method has been demonstrated empirically through a series of challenging cases. It is of future interest to improve the completeness of the formulated scenario state space and develop more sampling-efficient safe set quantification algorithms. The proposed method is also expected to generalize to the safety evaluation of other cooperative car-following systems and human drivers within the same operable domain.

\bibliography{ifacconf}             

\begin{thebibliography}{18}
\providecommand{\natexlab}[1]{#1}
\providecommand{\url}[1]{\texttt{#1}}
\providecommand{\urlprefix}{URL }
\expandafter\ifx\csname urlstyle\endcsname\relax
  \providecommand{\doi}[1]{doi:\discretionary{}{}{}#1}\else
  \providecommand{\doi}{doi:\discretionary{}{}{}\begingroup
  \urlstyle{rm}\Url}\fi

\bibitem[{EuroNCAP(2019)}]{ncapc2c2019}
EuroNCAP (2019).
\newblock European new car assessment programme (euro ncap) test protocol –
  {AEB} car-to-car systems.
\newblock Technical report, The European New Car Assessment Programme.

\bibitem[{Fan et~al.(2017)Fan, Qi, Mitra, and Viswanathan}]{fan2017d}
Fan, C., Qi, B., Mitra, S., and Viswanathan, M. (2017).
\newblock D ry vr: data-driven verification and compositional reasoning for
  automotive systems.
\newblock In \emph{International Conference on Computer Aided Verification},
  441--461. Springer.

\bibitem[{Feng et~al.(2020)Feng, Feng, Yan, Shen, Xu, and Liu}]{feng2020safety}
Feng, S., Feng, Y., Yan, X., Shen, S., Xu, S., and Liu, H.X. (2020).
\newblock Safety assessment of highly automated driving systems in test tracks:
  a new framework.
\newblock \emph{Accident Analysis \& Prevention}, 144, 105664.

\bibitem[{Forkenbrock and Snyder(2015)}]{forkenbrock2015nhtsa}
Forkenbrock, G.J. and Snyder, A.S. (2015).
\newblock {NHTSA}’s 2014 automatic emergency braking test track evaluations.
\newblock Technical report, National Highway Traffic Safety Administration.

\bibitem[{Rao et~al.(2019)Rao, Forkenbrock et~al.}]{rao2019test}
Rao, S.J., Forkenbrock, G.J., et~al. (2019).
\newblock Test procedures traffic jam assist test development considerations.
\newblock Technical report, United States. Department of Transportation.
  National Highway Traffic Safety Administration.

\bibitem[{Shihadeh et~al.(2018)}]{openpiotgithub}
Shihadeh, A. et~al. (2018).
\newblock openpilot.
\newblock \url{https://github.com/commaai/openpilot}.

\bibitem[{Treiber and Kesting(2013)}]{treiber2013traffic}
Treiber, M. and Kesting, A. (2013).
\newblock Traffic flow dynamics.
\newblock \emph{Traffic Flow Dynamics: Data, Models and Simulation,
  Springer-Verlag Berlin Heidelberg}.

\bibitem[{Wang et~al.(2021)Wang, Xie, Huang, and Liu}]{wang2021review}
Wang, C., Xie, Y., Huang, H., and Liu, P. (2021).
\newblock A review of surrogate safety measures and their applications in
  connected and automated vehicles safety modeling.
\newblock \emph{Accident Analysis \& Prevention}, 157, 106157.

\bibitem[{Weng(2021)}]{weng2021model}
Weng, B. (2021).
\newblock A class of model predictive safety performance metrics for driving
  behavior evaluation.
\newblock In \emph{2021 IEEE International Intelligent Transportation Systems
  Conference (ITSC)}, 180--187.
\newblock \doi{10.1109/ITSC48978.2021.9565013}.

\bibitem[{Weng(2022)}]{sdq_tools}
Weng, B. (2022).
\newblock {SDQ} tools.
\newblock \url{https://gitlab.com/Bobeye/sdq_tools}.

\bibitem[{Weng et~al.(2021{\natexlab{a}})Weng, Capito, Ozguner, and
  Redmill}]{weng2021finite}
Weng, B., Capito, L., Ozguner, U., and Redmill, K. (2021{\natexlab{a}}).
\newblock A finite-sampling, operational domain specific, and provably unbiased
  connected and automated vehicle safety metric.
\newblock \emph{arXiv preprint arXiv:2111.07769}.

\bibitem[{Weng et~al.(2021{\natexlab{b}})Weng, Capito, Ozguner, and
  Redmill}]{weng2021formal}
Weng, B., Capito, L., Ozguner, U., and Redmill, K. (2021{\natexlab{b}}).
\newblock A formal characterization of black-box system safety performance with
  scenario sampling.
\newblock \emph{IEEE Robotics and Automation Letters}.
\newblock \doi{10.1109/LRA.2021.3122517}.

\bibitem[{Weng et~al.(2021{\natexlab{c}})Weng, Capito~Ruiz, Ozguner, and
  Redmill}]{weng2021towards}
Weng, B., Capito~Ruiz, L.J., Ozguner, U., and Redmill, K. (2021{\natexlab{c}}).
\newblock Towards guaranteed safety assurance of automated driving systems with
  scenario sampling: An invariant set perspective.
\newblock \emph{IEEE Transactions on Intelligent Vehicles}.
\newblock \doi{10.1109/TIV.2021.3117049}.

\bibitem[{Wishart et~al.(2020)Wishart, Como, Elli, Russo, Weast, Altekar,
  James, and Chen}]{wishart2020driving}
Wishart, J., Como, S., Elli, M., Russo, B., Weast, J., Altekar, N., James, E.,
  and Chen, Y. (2020).
\newblock Driving safety performance assessment metrics for ads-equipped
  vehicles.
\newblock \emph{SAE Technical Paper}, 2(2020-01-1206).

\bibitem[{Zhao et~al.(2017)Zhao, Huang, Peng, Lam, and
  LeBlanc}]{zhao2017accelerated}
Zhao, D., Huang, X., Peng, H., Lam, H., and LeBlanc, D.J. (2017).
\newblock Accelerated evaluation of automated vehicles in car-following
  maneuvers.
\newblock \emph{IEEE Transactions on Intelligent Transportation Systems},
  19(3), 733--744.

\bibitem[{Zhao et~al.(2016)Zhao, Lam, Peng, Bao, LeBlanc, Nobukawa, and
  Pan}]{zhao2016accelerated}
Zhao, D., Lam, H., Peng, H., Bao, S., LeBlanc, D.J., Nobukawa, K., and Pan,
  C.S. (2016).
\newblock Accelerated evaluation of automated vehicles safety in lane-change
  scenarios based on importance sampling techniques.
\newblock In \emph{IEEE Transactions on Intelligent Transportation Systems},
  volume~18, 595--607. IEEE.

\bibitem[{Zhong et~al.(2021)Zhong, Hu, Guo, Zhang, Zhong, and
  Ray}]{zhong2021detecting}
Zhong, Z., Hu, Z., Guo, S., Zhang, X., Zhong, Z., and Ray, B. (2021).
\newblock Detecting safety problems of multi-sensor fusion in autonomous
  driving.
\newblock \emph{arXiv preprint arXiv:2109.06404}.

\bibitem[{Zhu(2022)}]{openpilot_carla}
Zhu, M. (2022).
\newblock Openpilot in {C}arla.
\newblock \url{https://github.com/pgchui/openpilot_in_carla}.

\end{thebibliography}
                                                   









\end{document}